\documentclass[runningheads]{llncs}
\usepackage{multirow}
\usepackage[T1]{fontenc}
\usepackage{graphicx,verbatim}
\begin{document}
\title{SurgVLA-Bench: Towards Evaluating Vision-Language-Action Models for Laparoscopic Surgical Robotics}
%



\author{Jiashuo Sun\inst{1,2}\thanks{These authors contributed equally to this work. \protect\\ \protect\llap{$^{\dagger}$ }Corresponding author.} \and
Yue He\inst{1,3}$^{\star}$ \and
Wenxuan Liu\inst{1,3} \and
Tao Mao\inst{1,3} \and
Jiazheng Wang\inst{1,3} \and
Xiang Chen\inst{1,3} \and
Min Liu\inst{1,2,3}$^{\dagger}$}

\authorrunning{J. Sun et al.}

\titlerunning{SurgVLA-Bench}

\institute{
the National Engineering Research Center of Robot Visual Perception and Control Technology, China \and
the national graduate college for elite engineers, Hunan University, Hunan, China \and
the School of Artificial Intelligence and Robotics, Hunan University, Hunan, China \\
\email{SunJiashuo@hnu.edu.cn,liu\_min@hnu.edu.cn}}

\maketitle              
\begin{abstract}
Vision-Language-Action (VLA) models represent a promising direction for embodied intelligence in surgical robotics. Despite the prevalence of VLA benchmarks for general robotics, standardized evaluation platforms specifically designed for surgical contexts remain absent. 
To address this limitation, we present SurgVLA-Bench, the first comprehensive benchmark for evaluating VLA models in laparoscopic surgical robotics. Leveraging the SurRoL simulation platform, we construct a hierarchical task taxonomy ranging from atomic actions to complete surgical procedures, complemented by a multi-dimensional evaluation framework assessing action accuracy and semantic consistency. We then systematically evaluate two representative paradigms, including autoregressive models such as OpenVLA, and flow matching models such as $\pi_{0}$, $\pi_{0.5}$, and SmolVLA. Our experiments show that autoregressive models tend to excel in semantic understanding, while flow matching models often achieve higher task precision but may face generalization trade-offs. However, even the best-performing models remain far from satisfactory, as the constrained endoscopic field of view, restricted viewing angles, and frequent occlusions persist as fundamental physical bottlenecks. The code and data are available at https://github.com/VCL-HNU/SurgVLA . 


\end{abstract}
\section{Introduction}

Surgical robots have become increasingly integral to modern clinical practice, offering enhanced precision, dexterity, and stability over conventional manual procedures. However, current systems still rely heavily on continuous teleoperation by experienced surgeons, motivating pursuit of higher autonomy levels. In this context, Vision-Language-Action (VLA) models \cite{zhang2025pure} have emerged as a promising paradigm, demonstrating remarkable capabilities in understanding language instructions, executing task-specific actions, and generalizing across diverse scenarios \cite{attanasio2021autonomy}. This raises a question: Can VLA models enable intelligent surgical robots to understand commands and perform precise interventions?

The field of VLA models has recently experienced significant advances. RT-2 \cite{rt2} pioneered the approach of using vision-language model output tokens directly as robot actions. OpenVLA \cite{openvla} introduced the first open-source VLA framework, providing architectural insights for subsequent model development. PaLM-E \cite{palm} demonstrated the feasibility of training VLA models on internet-scale data. $\pi_0$ \cite{pi-0} introduced flow matching \cite{lipman2022flow} with multi-stage training, showing positive effects on task generalization. SmolVLA \cite{smolvla} and VLA-Adapter \cite{vla-adapter} explored lightweight approaches for efficient VLA deployment, while RyNN-VLA \cite{rynnvla} combined embodied models with world models to enhance environmental understanding. However, these models are predominantly developed for general-purpose scenarios and lack applicability to surgical contexts, including the ability to discriminate surgical-specific objects, meet millimeter-level precision requirements, and address safety concerns arising from model hallucinations \cite{kim2025medical}.

Research on surgical autonomy has primarily relied on task-specific methods, including imitation learning for suturing and peg transfer~\cite{pore2021learning,soper2008fundamentals}, reinforcement learning in simulation~\cite{surrol,yu2024orbit} and on physical platforms~\cite{kim2024surgical}, and vision-language pretraining for surgical scene understanding~\cite{yuan2024hecvl}. While effective within their respective scopes, these approaches each address only a subset of the perception-language-action pipeline, and none provides a unified framework that jointly integrates visual understanding, language instruction following, and action generation. On the evaluation side, general-purpose robotics benchmarks such as LIBERO~\cite{liu2023libero}, CALVIN~\cite{mees2022calvin}, and RLBench~\cite{james2020rlbench} have been instrumental in driving VLA research by providing standardized task suites and reproducible evaluation protocols. In the surgical domain, datasets such as Cholec80~\cite{twinanda2016endonet} and CholecT50~\cite{nwoye2023cholectriplet2021} have advanced skill assessment and scene recognition, yet they focus on understanding rather than closed-loop action generation. No existing benchmark simultaneously evaluates visual perception, language instruction following, and action generation in surgical environments.

To bridge this gap, we propose SurgVLA-Bench, aiming to explore the capability of VLA models to understand surgeon instructions and execute precise interventions in surgical scenarios. This is the first VLA evaluation benchmark specifically designed for surgical contexts. Built upon the SurRoL \cite{surrol} platform, our benchmark encompasses a comprehensive suite of evaluations ranging from atomic action tasks to composite surgical procedure segments.

Our main contributions are summarized as follows:  
\begin{enumerate}
    \item We propose SurgVLA-Bench, the first benchmark for surgical VLA models, featuring a hierarchical task taxonomy from atomic actions to full procedures and built on simulated environments with surgical instruments and multi-organ anatomical structures.
    \item We construct a comprehensive standardized dataset supporting multiple mainstream formats (RLDS \cite{o2024open}, LeRobot \cite{lerobot}, etc.), facilitating direct adoption and extension by the research community.
    \item We benchmark four representative VLAs \cite{openvla,smolvla,pi0.5,pi-0}, analyzing their limitations in surgical contexts and identifying key challenges for future research.

\end{enumerate}

\section{Benchmarking}
\subsection{Tasks}
\begin{figure}[htb]
\begin{center}
\includegraphics[width=0.9\textwidth]{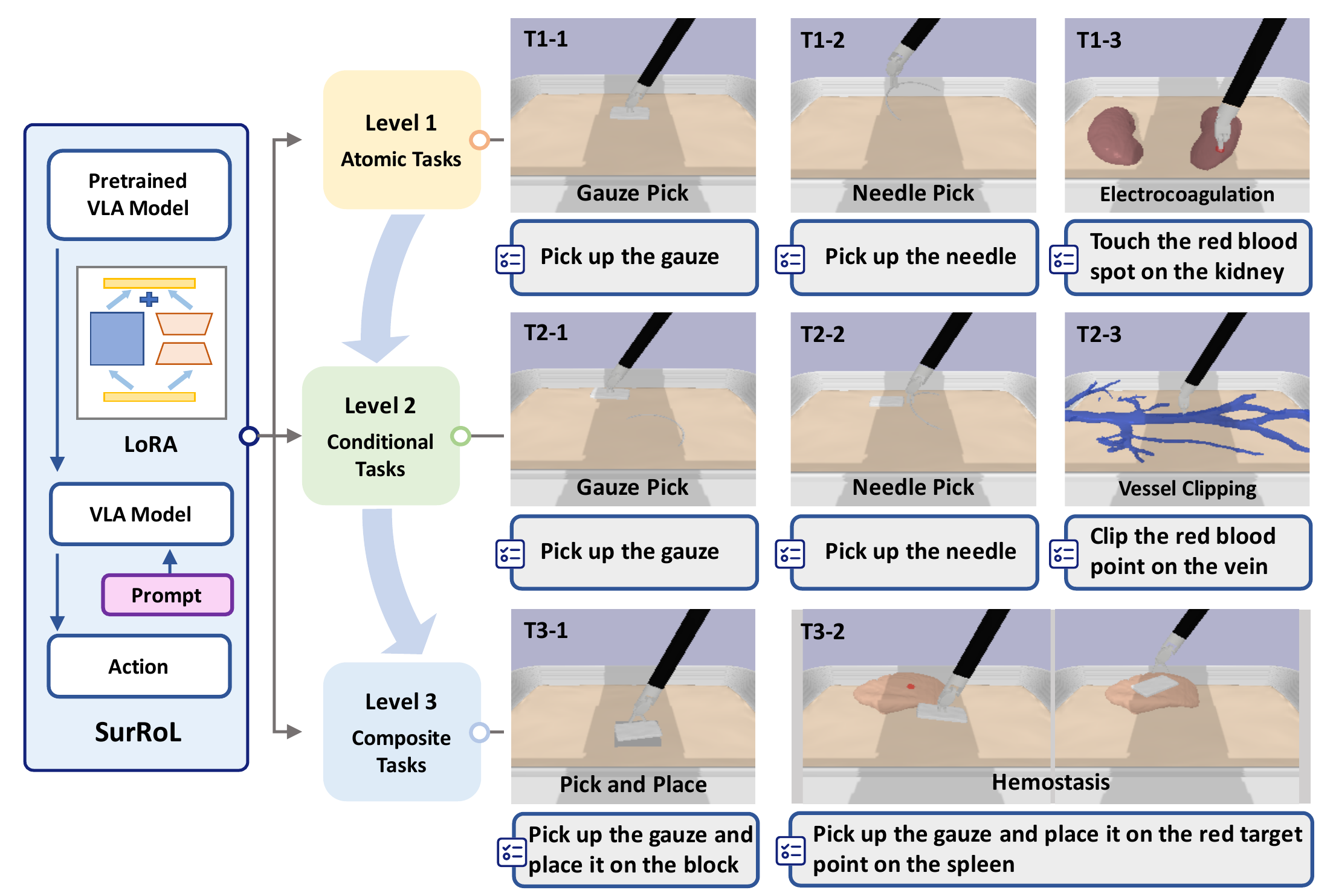}
\caption{Overview of SurgVLA-Bench. The left panel illustrates the fine-tuning and deployment pipeline, where a pretrained VLA model is fine-tuned via LoRA and deployed in the SurRoL simulation environment. The right panel presents the hierarchical task taxonomy spanning three difficulty levels: Atomic, Conditional, and Composite Tasks.
}
    \label{fig:overview}
\end{center}
\vspace{-2ex}
\end{figure}

Surgical procedures are inherently hierarchical: clinical workflows combine basic actions into complex operations, with different levels imposing distinct capability demands. To systematically evaluate VLA models in surgical scenarios and identify their capability bottlenecks, we design a clinically relevant hierarchical evaluation system encompassing eight tasks. Based on complexity, tasks are categorized into three levels (Fig. \ref{fig:overview}). The design follows three principles: (1) gradual difficulty progression; (2) coverage of typical surgical operations; (3) support for independent evaluation of different VLA capability dimensions.
Each level targets distinct capabilities: Level 1 focuses on action precision, Level 2 on target discrimination and instruction following, and Level 3 on multi-step planning and spatial localization.

\textbf{Level 1: Atomic Tasks.} The scene contains only one target object without distractors, and language instructions are direct and unambiguous. This level primarily evaluates action precision and basic control capabilities. It includes three tasks: Gauze Pick (T1-1) and Needle Pick (T1-2), which simulate basic grasping of surgical consumables, and Electrocoagulation (T1-3), which simulates contact-based cauterization of a target tissue site.

\textbf{Level 2: Conditional Tasks.} The scene contains distractors, or the target object itself presents multiple confounding factors, requiring the model to accurately identify the manipulation target based on language instructions. This level primarily evaluates target discrimination capability and instruction-following accuracy. It includes: Gauze Pick (T2-1) and Needle Pick (T2-2), which extend their Level 1 counterparts by introducing visually similar distractors into the scene, and Vessel Clipping (T2-3), which simulates the placement of a hemostatic clip on a designated vessel segment among multiple candidate sites.

\textbf{Level 3: Composite Tasks.} The model must complete comprehensive operational workflows, placing objects at designated target positions. This level primarily evaluates multi-step sequential execution capability and precise spatial localization ability. It includes: Pick and Place (T3-1), which requires grasping a target object and transferring it to a specified location, and Hemostasis (T3-2), which simulates a complete hemostatic procedure involving sequential gauze grasping, transport, and placement onto a bleeding site on an organ surface.

\subsection{Dataset Construction}

To support the evaluation of SurgVLA-Bench, we construct a dedicated surgical task dataset. Unlike general robotics, where large-scale open-source datasets such as Open X-Embodiment \cite{o2024open} are readily available for VLA model training and evaluation, the surgical domain currently lacks standardized datasets suitable for this purpose. Therefore, we collect and construct our dataset from scratch based on task scenarios within the SurRoL \cite{surrol} simulation environment.

\begin{figure}[htbp]
\begin{center}
\includegraphics[width=0.95\textwidth]{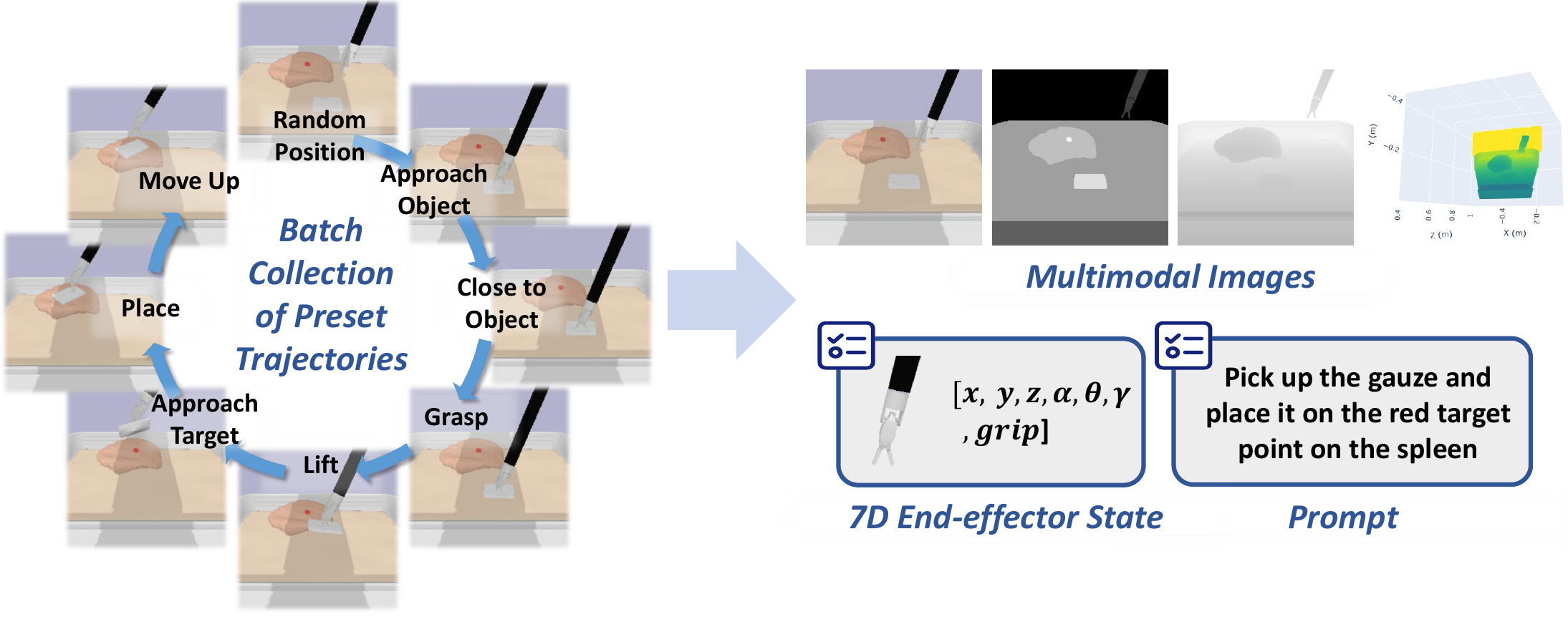}
\caption{Data Acquisition Pipeline and Dataset Composition.}
    \label{fig:dataset}
\end{center}
\vspace{-2ex}
\end{figure}

\textbf{Data Collection.}We collect trajectories for all eight tasks defined in Section 2.1, organized into three datasets corresponding to the three difficulty levels. Taking T3-2 as an example, the data collection process is illustrated in the Fig. \ref{fig:dataset}. We randomly position the target within a predefined area and execute the specified actions along preset trajectories based on the object's location. Each trajectory contains RGB images, depth images, robot proprioceptive states, and corresponding action sequences. For consistency, all RGB images are captured at a resolution of [3 × 224 × 224]. In total, we collect over 800 complete trajectories comprising approximately 40,000 action frames across all eight tasks, with each trajectory requiring manual verification to ensure action quality. Since the primary objective at this stage is to verify the basic feasibility of VLA models on the benchmark, we adopt concise and unambiguous instruction descriptions for each task (see Fig. \ref{fig:overview}), ensuring a clear correspondence between language instructions and task objectives. Additionally, we develop an integrated data collection pipeline that enables future developers to expand the dataset scale according to their needs. We provide the data in LeRobot \cite{lerobot}, RLDS \cite{o2024open}, and 3D point cloud formats to facilitate direct adoption across different VLA paradigms.

\textbf{Gripper State Representation.} To mitigate the sparsity of the original pulse-like gripper signal, we convert the gripper action from instantaneous change values to continuous state values (i.e., persistently outputting closing or opening signals), which significantly improves training efficiency.


\subsection{Simulation Environment}

We construct surgical simulation scenarios based on the SurRoL simulation platform. First, we integrate multiple high-fidelity anatomical organ models, including liver, kidney, and venous vasculature, and combine them with existing instrument models from the SurRoL platform to build diverse surgical scenes. To enhance anatomical realism, we ensure that all organ models maintain accurate anatomical proportions. Beyond the organ models used in task scenarios, we provide an additional library of abdominal cavity organ models covering major intra-abdominal organs, facilitating future extensions by developers.

Furthermore, SurRoL natively employs the PyBullet \cite{coumans2016pybullet} physics engine, whose contact detection is primarily designed for rigid bodies and has limited effectiveness for simulating deformable objects. So we optimize the grasping mechanism to address PyBullet's limited effectiveness for deformable objects. Informed by real-world experiments, we replace physics-based contact detection with a task-specific binding mechanism that triggers grasping when spatial and kinematic conditions derived from real-world observations are jointly satisfied, yielding manipulation outcomes that more faithfully reflect real surgical robot behavior.

\section{Experiments}

\subsection{Implementation Details}

\textbf{Model Selection.}
To comprehensively evaluate VLA model performance in surgical scenarios, 
We select four representative VLA models as baselines, covering two mainstream paradigms: OpenVLA~\cite{openvla}, built on Prismatic-7B~\cite{karamcheti2024prismaticvlmsinvestigatingdesign}, represents the autoregressive approach; the $\pi$-series ~\cite{pi0.5,pi-0} and SmolVLA~\cite{smolvla} ($\sim$0.5B parameters) adopt Flow Matching action heads, with $\pi$0.5 adding a dual-layer policy architecture. Architectural details are summarized in Table~\ref{tab:models}.


\begin{table}[htbp]
\begin{center}
\caption{
Architectural details of baseline VLA models evaluated in SurgVLA-Bench.
}\label{tab:models}
\small
\setlength{\tabcolsep}{4pt}
\begin{tabular}{ lccccccccc }
\hline
\hline
VLA Model & Year & Vision Encoder & LLM & Action Head\\
\hline
OpenVLA \cite{openvla} & 2024  & SigLIP + DINOv2  & Llama 2 7B & Autoregressive \\
$\pi_0$ \cite{pi-0} & 2024  & SigLIP 400M  & Gemma 2B & Flow Matching \\
$\pi_{0.5}$ \cite{pi0.5} & 2025  & SigLIP2  & Gemma 2 & Flow Matching \\
SmolVLA \cite{smolvla} & 2025 & SigLIP & SmolLM2 &  Flow Matching\\
\hline
\hline
\end{tabular}
\end{center}
\vspace{-2ex}
\end{table}

\textbf{Experimental Setup.}
As illustrated in Fig. \ref{fig:overview}, we design language instructions in a colloquial style for both atomic actions and tasks that simulate real surgical operations. The instructions are phrased in natural conversational language rather than strict medical terminology to facilitate LLM comprehension of surgical scenarios, thereby maximizing instruction following in VLA models.

All experiments are conducted on 4 NVIDIA A6000 GPUs. For each task, 50 independent trials were conducted per model. Following the evaluation protocols established in prior work \cite{openvla,smolvla,pi-0,pi0.5}, we report the resulting success rate (SR) as the primary evaluation metric. The evaluation employs a binary success/failure judgment under step limits without time constraints. The maximum allowable steps are set to 100 for Level 1 and Level 2 tasks, and 150 for Level 3 tasks. To ensure fair comparison, all models are fine-tuned using Low-rank adaptation (LoRA) \cite{hu2022lora} until convergence. Specifically, each model is trained separately on a per-level basis and subsequently evaluated on individual sub-tasks within either the same level or across different levels. All models are evaluated under identical simulation environments, task configurations, and evaluation protocols, allowing each model to fully demonstrate its performance potential.

\subsection{Experiments and Analysis}

\begin{table}[htbp]
\begin{center}
\caption{
Results on SurgVLA-Bench. For each evaluation task, best results per column are highlighted in bold.
}\label{tab:results}
\small
\setlength{\tabcolsep}{5.5pt}
\begin{tabular}{ lcccccccc }
\hline
\hline
\multirow{2}{*}{Model} & \multicolumn{3}{c}{Level 1} & \multicolumn{3}{c}{Level 2} & \multicolumn{2}{c}{Level 3}  \\ 
\cline{2-4} \cline{5-7} \cline{8-9}
  & T1-1 & T1-2 & T1-3 & T2-1 & T2-2 & T2-3 & T3-1 & T3-2   \\
\hline
OpenVLA \cite{openvla} & 36\% & 2\% & \textbf{76\%} & 8\% & 2\% & \textbf{72\%} & 0\% & 0\%  \\
\hline
$\pi_0$ \cite{pi-0} & 76\% & 0\% & 0\% & 0\% & 4\% & 6\% & \textbf{56\%} & \textbf{2\%}  \\
$\pi_{0.5}$ \cite{pi0.5} & \textbf{78\%} & \textbf{8\%} & 0\% & \textbf{36\%} & \textbf{10\%} & 14\% & 12\% & 0\% \\
SmolVLA \cite{smolvla} & 0\% & 8\% & 0\% & 0\% & 0\% & 10\% & 0\% & 0\%  \\
\hline
\hline
\end{tabular}
\end{center}
\end{table}

\textbf{Analysis of Success Rate.}
As shown in Table \ref{tab:results}, OpenVLA achieves competitive overall performance after fine-tuning. While its precision on atomic tasks is moderate, it effectively identifies surgical tools and anatomical structures, excelling on organ-involved tasks T1-3 and T2-3. OpenVLA localizes well but struggles with gripper closure, impairing gauze grasping. For needle grasping, the needle's slender, curved geometry hinders all models from learning correct gripper rotation, causing uniformly poor performance.

The $\pi$-series models handle single-object gauze grasping (T1-1) but fail on needle (T1-2) and electrocautery (T1-3) tasks, reflecting limited multi-task semantic discrimination, indicating limited multi-task semantic discrimination. This limitation is further evidenced by their degraded performance on Level 2 tasks involving multiple objects, where $\pi_{0.5}$ shows improvement over $\pi_0$. However, $\pi_0$ achieves strong results on the longer-horizon task T3-1 at Level 3, indicating its ability to execute multi-step tasks without distractions. In contrast, $\pi_{0.5}$ exhibits lower success rates on this task, likely due to its hybrid action representation combining discrete tokens for high-level semantic planning and continuous flow matching for low-level control. The lack of high-level supervision during fine-tuning may hinder its performance on multi-stage long-horizon tasks. Both $\pi$-series models perform poorly on organ-involved tasks, likely attributed to the domain gap between surgical scenes and pre-training data, where similar anatomical priors are absent.

As a lightweight model, SmolVLA struggles with scene understanding under multi-task settings with randomly placed objects and fails to learn proper gripper actions. Yet in simple scenarios like T1-1, it consistently reaches the correct position, with failure stemming only from gripper closure. Under single-task training, SmolVLA achieves 82\% SR on T1-1, demonstrating its potential in simplified settings.

\begin{figure}[t!]
\begin{center}
\includegraphics[width=\textwidth]{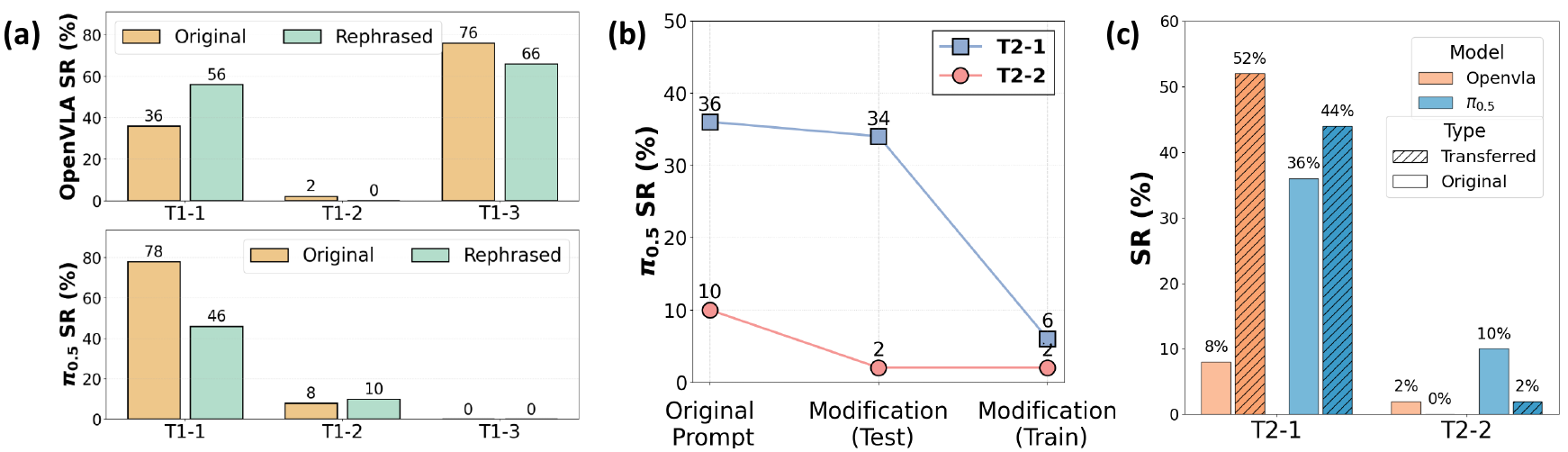}
\caption{Model analysis experiments. (a) Performance of OpenVLA and $\pi_{0.5}$ under prompt reformulation; (b) Impact of detailed prompts on $\pi_{0.5}$ during test-time and training-time; (c) Direct transfer results of OpenVLA and $\pi_{0.5}$ from Level 1 to Level 2.
}
    \label{fig:analysis}
    \vspace{-2ex}
\end{center}
\end{figure}

\textbf{Analysis of Prompt Robustness.}
To assess prompt ambiguity effects, we rephrased instructions for OpenVLA \cite{openvla} and $\pi_{0.5}$\cite{pi0.5} ("lift" instead of "pick up", "contact" instead of "touch"). As shown in Fig. \ref{fig:analysis}(a), OpenVLA exhibits limited performance degradation under instruction perturbations, with improved success rates on modified gauze grasping commands. In contrast, $\pi_{0.5}$ shows larger performance drops on task T1-1, indicating that the LLM of OpenVLA maintains superior robustness after fine-tuning.

\textbf{Analysis of Prompt Informativeness Utilization.}
To explore why the $\pi$-series models\cite{pi-0,pi0.5} struggle with multi-task generalization after LoRA fine-tuning, we test whether prompt modifications could help using two strategies: (1) using more detailed instructions at test time, and (2) expanding task descriptions during training (e.g., "pick up the gauze, not the needle"). As a more complex and higher-performing model than $\pi_0$, $\pi_{0.5}$ exhibits stronger coupling between its performance and architectural design. Consequently, LoRA fine-tuning disrupts its internal structure, degrading the LLM's instruction-following capability. As shown in Fig. \ref{fig:analysis}(b), modifying prompts for $\pi_{0.5}$ during testing or training not only fails to improve performance but leads to degradation, particularly when prompts are modified during training. The model favors simpler prompts over more informative but complex ones. This suggests that preserving pre-trained language capabilities under low-cost fine-tuning remains an open challenge for sophisticated VLA architectures \cite{li2025virt}.

\textbf{Analysis of Generalization.}
Both Level 1 and Level 2 contain gauze and needle objects, with Level 1 being simpler. To assess whether models learn object semantics, we directly apply models trained on level 1 to Level 2 tasks without fine-tuning, with results shown in Fig. \ref{fig:analysis}(c).
For T1-1 (gauze grasping), OpenVLA shows significant performance gain after transfer. This improvement likely stems from cleaner training signals in Level 1, which lacks interference from other objects, suggesting OpenVLA effectively learns semantic representations that transfer to more complex scenarios. $\pi_{0.5}$ maintains its performance after transfer, indicating it also captures gauze semantics.
For T2-2 (needle grasping), both models exhibit slight performance degradation, suggesting that direct transfer compromises spatial precision for fine-grained manipulation tasks.

\section{Conclusion}


In this paper, we introduce SurgVLA-Bench, the first benchmark for evaluating Vision-Language-Action models in surgical robotics. Built on SurRoL, our hierarchical evaluation system spans three levels and eight tasks with a dedicated dataset and multi-dimensional protocol. Systematic evaluation of four representative VLA models reveals that none perform satisfactorily under multi-task training and testing. Autoregressive models tend to excel at semantic understanding, whereas flow matching models often deliver higher precision in task execution. However, this strength in specificity may come at the cost of limited generalization. In surgical scenarios involving multiple complex sub-tasks, addressing training conflicts among different tasks remains a key direction for future VLA research. These limitations stem from task interference under multi-task training, inherent differences in robot kinematics, limited surgical data, and most critically, the constrained endoscopic view with restricted angles and occlusions that impede depth perception.
We believe SurgVLA-Bench can serve as a standardized evaluation platform for VLA model research in surgical scenarios and help drive further advancements in this field. Future work will extend the benchmark with greater task diversity, and real surgical data validation.

\bibliographystyle{splncs04} 
\bibliography{Paper-1457}

\end{document}